\documentclass[journal]{IEEEtran}
\usepackage{xcolor}
\usepackage{cite}
\usepackage{hyperref}
\usepackage{amsmath}
\usepackage{amssymb}
\usepackage{pifont}
\usepackage{multirow}
\usepackage{array}
\usepackage{footnote}

\usepackage{moreverb,url}

\usepackage{algorithm,algorithmic}%

\usepackage[pdftex]{graphicx}
\ifCLASSINFOpdf
\else
\fi
\usepackage{amsfonts}

\usepackage{stfloats}
\begin{document}
\title{An Attention-guided Multistream Feature Fusion Network for Localization of Risky Objects in Driving Videos}


\author{Muhammad Monjurul Karim,
        Ruwen Qin\IEEEauthorrefmark{1}, {\it Member},
        Zhaozheng Yin, {\it Senior Member}
 % <-this % stops a space
\thanks{Muhammad Monjurul Karim, and Ruwen Qin are with the Department
of Civil Engineering,  Stony  Brook  University, Stony Brook, NY 11794, USA.}% <-this % stops a space
\thanks{Zhaozheng Yin is with the Department of Biomedical Informatics, Department of Computer Science, and AI Institute, Stony Brook University, Stony Brook, NY 11794, USA.}% <-this % stops a space% 
\thanks{\IEEEauthorrefmark{1} Corresponding author: Ruwen Qin, email: ruwen.qin@stonybrook.edu}
\thanks{Manuscript received MM DD, YYYY; revised MM DD, YYYY.}}


\markboth{Journal of \LaTeX\ Class Files,~Vol.~XX, No.~X, MM~YYYY}%
{Shell \MakeLowercase{\textit{et al.}}: Bare Demo of IEEEtran.cls for IEEE Journals}



\maketitle

\begin{abstract}
Detecting dangerous traffic agents in videos captured by vehicle-mounted dashboard cameras (dashcams) is essential to facilitate safe navigation in a complex environment. Accident-related videos are just a minor portion of the driving video big data, and the transient pre-accident processes are highly dynamic and complex. Besides, risky and non-risky traffic agents can be similar in their appearance. These make risky object localization in the driving video particularly challenging. To this end, this paper proposes an attention-guided multistream feature fusion network (AM-Net) to localize dangerous traffic agents from dashcam videos. Two Gated Recurrent Unit (GRU) networks use object bounding box and optical flow features extracted from consecutive video frames to capture spatio-temporal cues for distinguishing dangerous traffic agents. An attention module coupled with the GRUs learns to attend to the traffic agents relevant to an accident. Fusing the two streams of features, AM-Net predicts the riskiness scores of traffic agents in the video. In supporting this study, the paper also introduces a benchmark dataset called Risky Object Localization (ROL). The dataset contains spatial, temporal, and categorical annotations with the accident, object, and scene-level attributes. The proposed AM-Net achieves a promising performance of 85.73\% AUC on the ROL dataset. Meanwhile, the AM-Net outperforms current state-of-the-art for video anomaly detection by 6.3\% AUC on the DoTA dataset. A thorough ablation study further reveals AM-Net's merits by evaluating the contributions of its different components.

\end{abstract}


% \fi
\IEEEpeerreviewmaketitle

\section{Introduction}

Autonomous driving and Advanced Driver Assistance Systems (ADAS) have attracted attention and made rapid progress in recent years \cite{eskandarian2019research}. Although research is going in a favorable direction with a beautiful vision of more comfortable and safer driving experiences, there are still concerns about traffic accidents. From 2014 to August 23, 2022, 506 autonomous vehicle collisions were reported in California \cite{ca_dmv_2022}. Moreover, according to the 2018 Global Status Report on Road Safety from World Health Organization, about 1.35 million people are killed in traffic accidents yearly \cite{world2018global}. Developing intelligent driving systems to help drivers (or autonomous systems) identify and localize potential risks is urgently needed to reduce collisions and fatalities.

%In recent years, autonomous driving and advanced driver assistance systems have attracted attention and made rapid progress \cite{eskandarian2019research}. Although research is going in a favorable direction with a beautiful vision of more comfortable and safer driving experience, there are still concerns about traffic accidents. From 2014 to August 23, 2022, 506 autonomous vehicle collisions were reported in California, USA \cite{ca_dmv_2022}. Moreover, according to the 2018 global status report on road safety from World Health Organization, about 1.35 million people are killed in traffic accidents every year \cite{world2018global}. To reduce collisions and fatality rates, developing intelligent driving systems to help drivers (or autonomous systems) identify and localize potential risks is urgently needed. 

% Widely available dashboard cameras are attracting researchers to develop camera-based solutions in attempt to fulfilling this challenging need because of their low cost and robust performance. 

Therefore, recognizing and localizing abnormalities in a driving video captured by vehicle-mounted dashboard cameras (dashcam), one type of low-cost and widely deployed sensors, is an essential task. Achieving this functionality can provide a valuable reference for the subsequent path decision of ADAS. Besides, detecting the presence of a risky object provides the driver a reference to avoid them. This capability applies to other applications, such as traffic safety, autonomous driving, and pedestrian protection \cite{alvi2020comprehensive,muhammad2020deep}.

%Therefore, recognizing, predicting, and localizing abnormalities in a natural driving video captured by vehicle mounted dashboard cameras is an essential task. Achieving this task can provide a valuable reference for the subsequent path decision of intelligent driving. Besides, detecting the presence of a risky object will allow the driver (or the controller in the case of autonomous vehicles) sufficient amount of time to avoid them. Moreover, this capability can be applied in a variety of applications such as traffic safety warning \cite{alvi2020comprehensive}, autonomous driving \cite{muhammad2020deep}, and pedestrian protection \cite{muhammad2020deep}.

Many computer vision studies have tackled a related task that detects anomalous events from a dashcam \cite{chan2016anticipating,bao2020uncertainty, bao2021drive,Karim2022_Dsta, karim2021toward, fang2022traffic}. That is, this stream of literature focuses on identifying frames where the risk of a traffic accident is present. Those studies did not localize where in the frame the risk is present or detect what traffic agents will involve in an accident. Localizing dangerous objects from a driving scene is critical for ADAS in the real world.

%Many works in computer vision has studied to achieve this task by utilizing deep learning techniques in order to detect risk or anomalous events from dashboard cameras \cite{chan2016anticipating,bao2020uncertainty, bao2021drive,Karim2022_Dsta, karim2021toward, fang2022traffic}. However, most of these methods focus on identifying frames in which the risk may be present, but do not attempt to localize where in the frame the risk is present or to detect which road participants will involve in an accident, which is very critical for real-world driving assistance systems.

Early prediction and localization of dangerous objects, which are likely to involve a future accident, from a highly dynamic and complex driving scene have three crucial challenges. First, the visual appearance of a traffic agent in the driving scene may not tell much about its risk level because of the very similar visual appearance of different traffic agents. Second, the time window for recognizing dangerous objects is short. Lastly, the long-term temporal dependency of traffic agents underlying the accident risk is hard to capture. Several attempts have explored this topic, mainly using deep learning methods \cite{gao2019goal,kozuka2017risky,zeng2017agent,li2020make,liu2018future,yao2022dota}. While these studies have laid a solid foundation for risky object localization, the opportunity to further improve the performance by extracting rich information from RGB videos is worth an exploration, especially when the training dataset is limited.

%Early prediction and localization of a risk in a highly dynamic and complex driving scenes have three key but difficult challenges: i) very similar visual appearance that frequently correspond to vastly different levels of risk and no risk, ii) very short temporal window to realize the spatio-temporal relationships of traffic agents for a risk,   iii) capturing long term temporal dependencies that underlie the risk. Although, there is not much prior work on the task of predicting and localizing risk from visual inputs taken from car-mounted cameras, a few attempts were made in this direction. Where the problem is approached with different deep learning methodologies. For example, \cite{ gao2019goal,kozuka2017risky} used supervised or weakly supervised learning algorithm by formulating the problem as a two class classification problem to learn the interactions among different traffic agents to localize risky regions from the RGB input. \cite{zeng2017agent, li2020make} learns spatio-temporal relationships of different traffic agents to produce risk scores for different traffic agents by taking appearance feature of those objects. Another approach is to reconstruct future frame or predict future trajectory of different objects to compare the inconsistency with the normal driving scenes to find the anomaly \cite{liu2018future,yao2022dota}. All these approaches have laid a solid foundation in risk localization, however, these approaches lack a method to extract and utilize rich information from the input video sequence to improve the performance with small training dataset.

In filling the gap, this paper proposes to build a deep learning framework for risky object localization from dashcam videos. Several research questions below motivate the design of the proposed framework. What information, both at the frame and object levels, can be extracted from the dashcam captured video, and do they contain essential cues for risky object localization? How are spatio-temporal patterns of traffic agents learned from the extracted information, and can the patterns effectively distinguish traffic agents by their risk level? Should specific traffic agents receive more attention than others in the driving scene? If that is the case, what attention mechanism to develop?

%To fill the gap, this paper proposes a framework to risky object localization. The proposed framework learns the spatio-temporal relationship of the traffic agents. To utilize rich information from the video sequence this study developed an attention guided multi-stream feature fusion network that learns risk relevant cues by considering spatial and temporal relations among different objects from their optical flow and trajectory feature as well as their historical memory. For the trajectory feature this study uses the bounding box information as a feature vector to feed as a separate input. This study found that this bounding box information plays very important role in model performance since evolution of bounding box over time holds very obvious cues of risky object. The features of each of the objects and the full frame are encoded into a hidden distributed representation using Gated Recurrent Unit (GRU) Network. Since importance of all the spatially distributed objects are not equal to realize a risk, a self-attention mechanism is developed to distribute attention weights to the hidden representation to provide attention to the most semantically important objects and their historical memory. This hidden representation provide learnable patterns to indicate how the objects evolve and turn into a risk.

The proposed study faces another difficulty. Existing related datasets miss one or a few pieces of essential information required for developing the proposed framework to localize dangerous objects in a driving scene. Therefore, this paper introduces a new dataset, called the Risky Object Localization (ROL), to address the issue of lacking a suitable dataset. ROL contains 1,000 accident videos with diverse risky objects in various driving scenarios. It provides comprehensive annotations at the object, accident, and scene levels. The annotations in ROL cover the interests of multiple communities, including intelligent transportation systems, traffic safety, computer vision, and deep learning. Thus, it not only addresses the need of this study but will accelerate convergence research like  \cite{LI2021105962}  at the interface of those disciplines.

%This paper also introduces a new dataset called Risky Object Localization (ROL) Dataset for localizing risky objects in traffic scene. ROL contains 1,000 accident videos with different risky objects in various driving scenarios. In addition to the object level risk annotation, ROL includes 9 different labels \cite{LI2021105962} such as accident beginning time, manner of collision, road type, related to roadway, and so on, which could support traffic accident avoidance research. Existing datasets do not contain spatial location annotation of the risky objects or do not annotate the beginning time of an accident. Therefore, those datasets cannot be used for evaluating the accuracy of risky object localization in a video frame or cannot evaluate the earliness of risk prediction.

In summary, this paper makes two contributions by introducing a new attention-guided multistream network (AM-Net) and a new dataset called ROL. As a result, the localization of risky traffic agents from the dashcam-captured driving scene achieves a new state-of-the-art performance. The remainder of this paper will further detail the discussion by presenting the following contents in sequence. Section \ref{sec:The Literature} summarizes the literature to determine the state-of-the-art. After that, Section \ref{sec:Methodology} delineates the proposed AM-Net, followed by the development of the ROL dataset in Section \ref{sec:Dataset Development}. Then, Section \ref{sec:results and Discussion} presents experimental studies to demonstrate and verify the merits of the contributed network and dataset. In the end, Section \ref{sec:Conclusion} summarizes research findings and important future work.

\section{The Literature}
\label{sec:The Literature}
This paper is built on studies that contribute to risky object localization, either directly or indirectly. The related literature is summarized below.  

\subsection{Anomaly Detection in a Video}

A topic related to risky object localization in a driving scene is video anomaly detection, which is about finding abnormal events in the video. Video anomaly detection is often formulated by profiling the normal behavior and measuring the spatial-temporal feature consistency. Deep learning-based methods can create a better normal video profile to determine video anomaly by measuring the consistency. For example, Hasan et al. \cite{hasan2016learning} developed a 3D Convolutional feed-forward Auto-Encoder (ConvAE) to model regular video frames. Motivated by this, Chong et al. \cite{chong2017abnormal} used a Convolutional Long Short Term Memory Auto-Encoder (ConvLSTMAE) to simultaneously take advantage of both Convolutional Neural Network (CNN) and Recurrent Neural Network (RNN) in modeling normal appearance and motion patterns. Liu et al. \cite{liu2018future} developed a Future Frame Prediction (FFP) method to generate future frames using past observations. At testing, observed deviation against the predicted content advises for abnormality. In these methods, normal situations are usually stable scenes, which limits their applicability in driving scenarios where the camera is in rapid motion.

%A topic closely linked to risky object localization in a driving scene is video anomaly detection, that is, finding abnormal events in videos. As for this field, the formulations of video anomaly detection are often defined by profiling the normal behavior and measuring the spatial-temporal feature consistency. Deep learning-based methods can create a better normal video profile to determine video anomaly by measuring the consistency. For example, Hasan et al. \cite{hasan2016learning} developed a 3D convolutional feed-forward autoencoder (Conv-AE) to model the regular video frames. Motivated by this, Chong et al. \cite{chong2017abnormal} used a convolutional LSTM Auto-Encoder to take advantage of both CNN and RNN to model normal appearance and motion patterns at the same time. Liu et al. \cite{liu2018future} developed a future frame prediction (FFP) method to generate future frames using past observation. At test time, observed deviation against the predicted content advise for abnormality. These methods need a stable scene motion for normal situations, which limits the applicability of these methods in practical driving scenarios where the camera is in rapid motion. 

%Unlike the above-mentioned static camera-based method, a few studies use mobile cameras to formulate the problem as specific types of anomalies, such as, abnormal motion \cite{singh2020anomalous}, pedestrians' action \cite{cai2021pedestrian}, risky lane change \cite{yurtsever2019risky}. However, these methods are very specific to an individual problem. 

\subsection{Traffic Accident Anticipation Using Dashcam Videos}

Traffic accident anticipation in dashcam videos has become a research focus recently. Unlike video surveillance systems, dashcam videos capture moving traffic agents that not only rapidly move but appear and disappear quickly in the scene. Different advanced methods are developed to learn the spatio-temporal pattern of the traffic agents to provide an overall riskiness score of the scene, including an LSTM predictor \cite{chan2016anticipating}, reinforced learning \cite{bao2021drive}, Graph Neural Network \cite{bao2020uncertainty, fang2022traffic}, and dynamic attention \cite{Karim2022_Dsta}. Although they only predict a risky event in the temporal domain, these studies have developed a solid methodological foundation for risky object localization in the spatial domain.

%Compared with anomaly detection in surveillance videos, traffic accident detection in dashcam videos has begun to be focused recently. Different from the video surveillance systems, the dashcam videos captured in moving cars are full of rapid motion and fast in and out of road participants. Different advanced methods using LSTM predictor \cite{chan2016anticipating}, reinforced learning \cite{bao2021drive}, Graph Neural Network \cite{bao2020uncertainty, fang2022traffic}, dynamic attention \cite{Karim2022_Dsta}, have been developed to learn the spatio-temporal pattern of the traffic agents to provide overall riskiness score of the scene. Although these methods only predict a risky event in temporal domain, they developed a strong foundation for risky object localization in spatial domain. 

\subsection{Risky Object Localization in a Driving Scene}

A few studies have attempted to solve the problem of risky object localization. Ohn-Bar et al. \cite{ohn2017all} developed a deep spatio-temporal importance prediction model that assigns riskiness scores to objects in a driving scene. Kozuka et al. \cite{kozuka2017risky} developed a weakly supervised method to predict pedestrian-involved risky regions. Zeng et al. \cite{zeng2017agent} proposed a soft-attention mechanism to provide agent-centric riskiness scores for different traffic agents. An RNN explicitly models non-linear interactions between agents. Although impressive, their methods did not utilize motion and trajectory features of the input video sequences, which are critical for risky object localization. Besides, limiting the number of candidate objects in each frame is not practical in many driving situations. Li et al. \cite{li2020make} studied the causal impact of risky objects to the driver’s behavior by removing candidate objects from the input video stream. While straightforward, it is a highly complex problem to solve in real-world applications because there could be many known and unknown casualties. Very recently, Yao et al. \cite{yao2022dota} proposed to localize risky objects by predicting future trajectory of traffic participants over a short horizon. Inconsistent predictions indicate that an anomaly has occurred or is about to happen. Although achieved a promising result, this method faces a new challenge arising from objects appearing suddenly in the scene. Those objects have limited information for predicting their future trajectories reliably.

\subsection{Risky Object Localization Datasets}

Several large-scale datasets emerged in response to the growing attention to deep learning-based risky event anticipation in traffic videos. Chan et al. \cite{chan2016anticipating} created the Dashcam Accident Dataset (DAD) by collecting 620 video clips of on-road accidents. The last ten frames of each video clip contain an accident. Yao et al. \cite{yao2019egocentric} developed the A3D dataset of 1,500 videos with annotations of accidents' starting and ending times. Bao et al. \cite{bao2020uncertainty} collected a car accident dataset of 1,500 videos to support traffic accident anticipation, which also contains annotations of environmental attributes and accident causes. These datasets are not directly applicable to the problem of risky object localization due to the lack of object-level risk annotation. Recently, Yao et al. \cite{yao2022dota} collected 4,600 videos with the risky object annotation. They annotated the risky objects contributing to an accident with their bounding boxes in videos when annotators judge the accident seems inevitable. This annotation approach is subjective. Besides, their annotations do not include the beginning time of accidents, which is necessary information to assess the earliness of a model's risky object localization ability.

\section{Methodology}
\label{sec:Methodology}
% \subsection{Framework overview}

\begin{figure*}[htbp]
\centering
\includegraphics[width=6.5in]{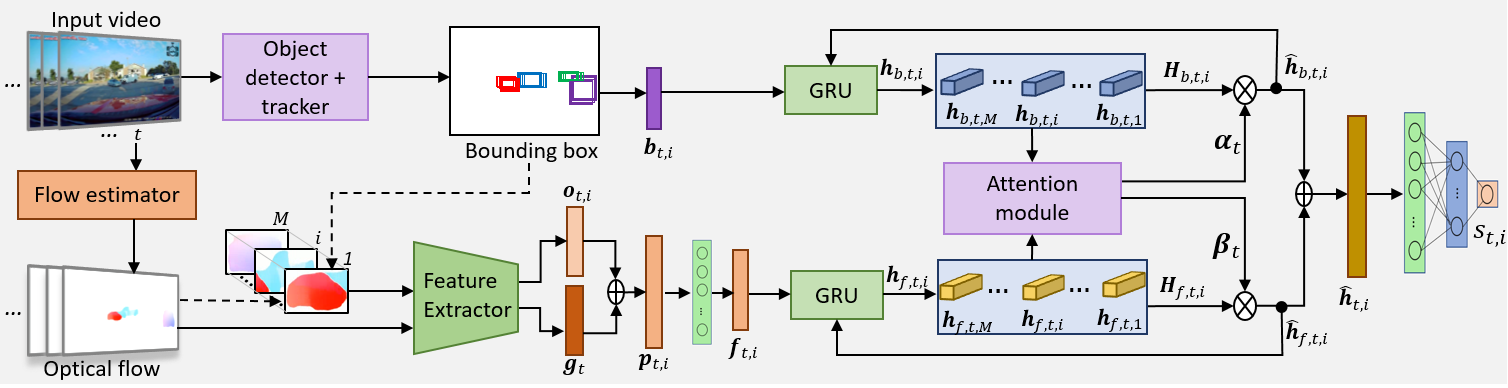}
\caption{Overview of the proposed AM-Net framework}
\label{fig_overview}
\end{figure*}

In filling the gap in the current studies, this paper proposes an Attention-Guided Multistream Network (AM-Net) in Fig. \ref{fig_overview}. The model reads frames of an input video in their time sequence to predict the riskiness score of any object in each frame. The model output is the riskiness score of any object (indexed as $i$) detected in each frame (indexed as $t$), $s_{t,i}$, for $t=1, 2, ...$ and any $i$. Traffic accidents usually involve collisions of different traffic agents. Sudden visual changes in the driving scene, erratic movement, and chaotic changes in the spatial-temporal relationship between traffic agents usually indicate that a traffic accident may occur. Therefore, this study models the risky object localization using multiple features such as object motion, dynamics of object bounding boxes, and a short history of the objects in the driving scene. Details of the model are below.

%In driving scenarios, traffic accidents usually involve collisions of different road participants or traffic agents. In general, there could be sudden visual scene changes, inconsistent movement, and chaotic change in spatial-temporal relationships between traffic participants that can distinguish risky traffic agents from normal non risky agents. Therefore, this study models the risky object localization by considering multiple features such as, motion consistency, evolution of object bounding boxes, and historical memory. The proposed method developed an attention guided multi-stream network (AM-Net) using Gated recurrent unit. AM-Net aims to predict a riskiness score $s_{t,i}$ for all the traffic agents indexed by $i$ in a driving scene at time frame $t$. This score indicates the object's probability of being risky in the future frame. Figure \ref{fig_overview} demonstrates the framework of the proposed AM-Net. 

% Unlike other existing works the proposed method does not need to limit the number of traffic agents in a scene.

%%%%%%%%%%%%%%%%%%%%%%%%%%%%%%%%%%%%%%%%%%%%%%%%%%%%%%%%%%%%%%%%%%%%%%%%%%%%%%%%%
%Figure: Overview
%%%%%%%%%%%%%%%%%%%%%%%%%%%%%%%%%%%%%%%%%%%%%%%%%%%%%%%%%%%%%%%%%%%%%%%%%%%%%%%%%
%%%%%%%%%%%%%%%%%%%%%%%%%%%%%%%%%%%%%%%%%%%%%%%%%%%%%%%%%%%%%%%%%%%%%%%%%%%%%%%%%

\subsection{Feature Extraction and Aggregation}
\label{subec:Feature Extraction and Aggregation}

AM-Net uses a pretrained object detector YOLOv5 \cite{ge2021yolox} to detect traffic agents in each frame and provide the bounding boxes of the detected objects in the frame. $M$ is the number of detected objects, which may vary from one frame to another. Because the temporal association of the same object in successive frames is critical information for risky object localization, a multi-object tracker DeepSort \cite{wojke2017simple} is used to make the association of objects in different frames.  

%further generates tracklets that each is a short track of a moving object in the driving scene. 

%First, the network reads a dashcam video. Each frame of the video flows into an object detector to get a set of object bounding boxes, ${\{b_{t,i}}\}_{i=0}^M$, where, $i$ is the index of the objects and $M$ is the number of objects which is a variable. To fulfill the task of risky object localization, temporal object associations in successive frames are needed. Therefore, a multi-object tracker associates the tracklets of detected object bounding boxes.

The change in pixel-level motion among video frames is an important clue to find objects with unusual movement. Therefore, AM-Net uses a pre-trained RAFT model  \cite{teed2020raft} to extract the optical flow image for every frame. Then, regions in the flow image, determined by the detected objects' bounding boxes, are cropped to become object-level flow images. Then, the feature extractor ResNet50 \cite{he2016deep} turns the frame-level flow image into a feature vector, $\boldsymbol{g}_t(\in\mathbb{R}^{D})$, and any of those at the object level into another feature vector, $\boldsymbol{o}_{t,i}(\in\mathbb{R}^{D})$. Here, $D$ is the dimension of those feature vectors. Then, $\boldsymbol{o}_{t,i}$ is concatenated with $\boldsymbol{g}_{t}$ to become the overall flow feature vector for the $i$th object, $\boldsymbol{p}_{t,i}(\in\mathbb{R}^{2D})$:

\begin{equation}
    \boldsymbol{p}_{t,i}=[\boldsymbol{o}_{t,i}; \boldsymbol{g}_{t}],
\end{equation}
which captures the object's motion feature in the driving scene.
% After passing two fully connected layers, $\pmb{P}_{t,i}$ is flatten and reduced to the feature vector, $\boldsymbol{F}_{t,i}(\in\mathbb{R}^d)$:
{After further passing a fully connected layer $\phi$, the flow feature vector $\pmb{p}_{t,i}$ becomes a lower dimensional feature vector, $\pmb{f}_{t,i}\in(\mathbb{R}^{2d})$: }
%\textcolor{red}{Feature vector $\pmb{P}_{t,i}$ is embedded into a lower dimensional space with a fully connected layer to get a lower dimensional feature vector, $\boldsymbol{F}_{t,i}(\in\mathbb{R}^{2d})$}
\begin{equation}
    \pmb{f}_{t,i} = \phi(\pmb{p}_{t,i};\pmb{\theta}_{0}),
\label{eq:flow}
\end{equation}
where $\pmb{\theta}_0$ are learnable parameters of the fully connected layer.

%{where $\phi$ represents a fully connected layer and $\pmb{\theta}$ ($\in \mathbb{R}^{(2D+1)\times 2d}$) are learnable parameters of the fully connected layers.}

Bounding boxes' location and scale changes in successive frames capture the spatial dynamics of traffic agents over time. Therefore, for any object $i$ in frame $t$, its bounding box's location $(x_{t,i},y_{t,i})$ and its scale in width $w_{t,i}$ and height $h_{t,i}$ are encoded as a feature vector $\pmb{b}_{t,i}(\in\mathbb{R}^4)$:
\begin{equation}
    \pmb{b}_{t,i}=[x_{t,i};y_{t,i};w_{t,i};h_{t,i}].
\end{equation}

\subsection{Spatio-temporal Relational learning with GRUs}
\label{subsec:Spatio-temporal Relational learning with GRUs}

Two Gated Recurrent Units (GRUs) respectively encode the extracted bounding box feature and flow feature of any detected object into their hidden representations and update them over time. Fig. \ref{fig_overview} shows that, one GRU takes the bounding box feature vector of object $i$ in frame $t$, $\pmb{b}_{t,i}$, and the weighted hidden representation of the same object in the last frame, $\widehat{\pmb{h}}_{b,t-1,i}$, to update its hidden representation:
\begin{equation}
    \pmb{h}_{b,t,i} = GRU(\pmb{b}_{t,i}, \widehat{\pmb{h}}_{b,t-1,i}; \pmb{\theta}_1),
\label{eq:gru_bbox}
\end{equation}
where $\pmb{\theta}_1$ are learnable parameters of the GRU, and $\pmb{h}_{b,t,i}\in\mathbb{R}^n$, where $n$ is the number of hidden states of the GRU. In parallel, the second GRU takes the flow feature vector of the object, $\pmb{f}_{t,i}$, and its weighted hidden representation in the past frame, $\widehat{\pmb{h}}_{f,t-1,i}$, to update the hidden representation:
\begin{equation}
    \pmb{h}_{f,t,i} = GRU(\pmb{f}_{t,i}, \widehat{\pmb{h}}_{f,t-1,i}; \pmb{\theta}_2),
\label{eq:gru_flow}
\end{equation}
where $\pmb{\theta}_2$ are the learnable parameters, and $\pmb{h}_{f,t,i}\in\mathbb{R}^N$,  where $N$ is the number of hidden states of the GRU. The weighted hidden representations in eqs. (\ref{eq:gru_bbox}) and (\ref{eq:gru_flow}) will be introduced below.

\subsection{Attention Module}
\label{subsec:Attention module}

Objects in the driving scene have unequal importance for predicting the risk of traffic accident. Therefore, learnable attentions should be distributed among the detected objects. Denote $\pmb{H}_{b,t} \in\mathbb{R}^{n \times M}$ as the hidden representations of the $M$ objects' bounding box features in frame $t$:

%All the objects in a video frame are not equally important for predicting a risk. Therefore, learnable attentions should be distributed among the objects features. This study modeled to provide attention weights to the hidden representation of all the object's features. Denote $\pmb{H}^F_t \in\mathbb{R}^{d \times M}$ as the hidden representations of flow features of the $M$ objects at frame $t$:
\begin{equation}
    \boldsymbol{H}_{b,t}=[\boldsymbol{h}_{b,t,1}, \dots,  \boldsymbol{h}_{b,t,i} ,\dots, \boldsymbol{h}_{b,t,M}].
\end{equation}
% Here, $M$ is the highest number of objects that can be present in a frame.
Here, the number of detected objects $M$ is a variable that can vary over time. Since the number of objects is not a fixed number the spatio-temporal relational learning is not biased to unrelated features.
 
%Unlike other's work, this study did not fix the number of objects in a frame. In this study $M$ is a variable, which is the number of objects in a frame that can vary frame to frame. Therefore, in this work spatio-temporal relational learning is not biased with any unnecessary feature.
% Unlike others' works, this study did not randomly select bounding boxes when number of detected objects ($k$) is less than $M$ to make $k=M$.
% If the number of detected object bounding boxes ($k$) is less than $M$ then $M - k$ non objects features are padded with zero in this study. Therefore, spatio-temporal relational learning is not biased with any random feature. $M$ is selected in this study using the prior knowledge of the dataset in such a way that no object is missed in any frame.

The dynamic spatial attention weights for the hidden representations of bounding box features, $\boldsymbol{\alpha}_{b,t}(\in\mathbb{R}^{M})$, are computed as

%The spatial attention weights,  $\boldsymbol{\beta}_{t}\in\mathbb{R}^{M}$, are computed as
\begin{equation}
\boldsymbol{\alpha}_{b,t} = \gamma(\tanh(\boldsymbol{H}_{b,t}^{\text{T}})\boldsymbol{w}_b),
\label{eq:attention}
\end{equation}
where $\boldsymbol{w}_b(\in\mathbb{R}^{n})$ are learnable parameters, and $\gamma$ represents the softmax operation. Then, $\boldsymbol{\alpha}_{b,t}$ is used to turn $\boldsymbol{H}_{b,t}$ into a weighted aggregation, $\widehat{\pmb{H}}_{b,t}(\in \mathbb{R}^{n \times M})$:
\begin{equation}
\pmb{\hat{H}}_{b,t}=[\widehat{\pmb{h}}_{b,t,1},\dots,\widehat{\pmb{h}}_{b,t,i},\dots,\widehat{\pmb{h}}_{b,t,M}]=\boldsymbol{\alpha}_{b,t}^{\text{T}} \odot \boldsymbol{H}_{b,t}.
\label{eq:tenergy}
\end{equation}

%Therefore, Weighted aggregated $\pmb{\hat{H}}^F_{t}$ can be represented as:
%\begin{equation}
%    \boldsymbol{\hat{H}}^F_{t}=[\boldsymbol{\hat{h}}^F_{0}, \dots,  \boldsymbol{\hat{h}}^F_{i} ,\dots, \boldsymbol{\hat{h}}^F_{M}]_t.
%\end{equation}

The same attention mechanism is applied to the hidden representations of flow features,
\begin{equation}
    \pmb{H}_{f,t}=[\pmb{h}_{f,t,1},\dots,\pmb{h}_{f,t,i},\dots,\pmb{h}_{f,t,M}].
\end{equation}
to get their flow feature attention weights $\pmb{\alpha}_{f,t}(\in\mathbb{R}^M)$:
\begin{equation}
\boldsymbol{\alpha}_{f,t} = \gamma(\tanh(\boldsymbol{H}_{f,t}^{\text{T}})\boldsymbol{w}_f) ,
\end{equation}
where $\boldsymbol{w}_f(\in\mathbb{R}^N)$ are learnable parameters of the attention module and $\gamma$ is the softmax operation. $\pmb{\alpha}_{f,t}$ is applied to $\pmb{H}_{f,t}$ to obtain the weighted flow hidden representations:
\begin{equation}
\pmb{\hat{H}}_{f,t}=[\widehat{\pmb{h}}_{f,t,1},\dots,\widehat{\pmb{h}}_{f,t,i},\dots,\widehat{\pmb{h}}_{f,t,M}]=\boldsymbol{\alpha}_{f,t}^{\text{T}} \odot \boldsymbol{H}_{f,t}.
\label{eq:tenergy}
\end{equation}

%The same attention mechanism is also applied to the bounding box feature to get the attention weight $\boldsymbol{\alpha}_{t}\in\mathbb{R}^{M}$ with a learnable parameter $\boldsymbol{W}_{Ba}\in\mathbb{R}^{n}$. The weight is then applied to the aggregated bounding box hidden representations $\boldsymbol{H}^B_{t}$ to get the weighted aggregated hidden representations $\boldsymbol{\hat{H}}^B_{t}$ for the bounding box feature. 

\subsection{Riskiness Score Prediction}
\label{subsec:Riskiness Score Prediction}

As Fig. \ref{fig_overview} shows, for any object $i$, the two attention-weighted hidden representations, $\widehat{\pmb{h}}_{b,t,i}$ and $\widehat{\pmb{h}}_{f,t,i}$, will respectively flow into the corresponding GRUs to update the object's hidden representations in the next frame $t+1$, as discussed in Section \ref{subsec:Spatio-temporal Relational learning with GRUs}. Meanwhile, they are concatenated to become the overall hidden representation of the object, $\widehat{\pmb{h}}_{t,i}(\in\mathbb{R}^{n+N})$:
\begin{equation}
    \widehat{\boldsymbol{h}}_{t,i}=[\widehat{\pmb{h}}_{b,t,i};\widehat{\pmb{h}}_{f,t,i}].
\end{equation}

$\widehat{\pmb{h}}_{t,i}$ is decoded by two fully-connected layers $\phi$ to output the scores of positive and negative classes, which are further normalized by the softmax operation $\gamma$ to find the riskiness score of object $i$ in frame $t$, $s_{t,i}$,
\begin{equation} 
    {s_{t,i} = \gamma(\phi(\phi(\widehat{\pmb{h}}_{t,i};\pmb{\theta}_3);\pmb{\theta}_4)}, 
\label{eq:score}
\end{equation}
which is the probability that the object will be involved in an accident soon.

%Then, for the $i$th object, $\boldsymbol{\hat{h}}^F_{i}$ is concatenated with $\boldsymbol{\hat{h}}^B_{i}$, 
%\begin{equation}
%    \boldsymbol{h}'_{t,i}=[\boldsymbol{\hat{h}}^B_{t,i};\boldsymbol{\hat{h}}^F_{t,i}],
%\end{equation}
%to form the overall representation for that object, where, $\boldsymbol{h}'_{t,i}\in\mathbb{R}^{n+N}$.

%Concatenated hidden representations $\pmb{h}'_{t,i}$ are decoded by two fully-connected layers to output the scores of positive and negative classes. The scores are further normalized by a softmax operator to find the riskiness score of both classes. The riskiness score $s_{t,i}$ for each object $i$ at time frame $t$ can be expressed as:

% where $\pmb{\theta}$'s ($\in \mathbb{R}^{d\times d}$) are learnable parameters of the fully connected layers.
%\textcolor{red}{where $\pmb{\theta}_3$ ($\in \mathbb{R}^{((N+n)+1)\times k_1}$) and $\pmb{\theta}_4$ ($\in \mathbb{R}^{(k_1+1)\times k_2}$) are the learnable parameters of the fully connected layers. $k_1$ and $k_2$ are the number of nodes in each layer, respectively.}

\subsection{The Loss Function}

The goal of the model training process is to fit all the learnable parameters $\pmb{\theta}$'s and $\pmb{w}$'s of the proposed model in Sections \ref{subec:Feature Extraction and Aggregation}$\sim$\ref{subsec:Riskiness Score Prediction} by backpropagating a differentiable loss function. A training dataset contains videos that each has $T$ frames in total and $M$ objects in each frame. In frame $t$, the risk class of the $i$th object, is indicated by the ground truth label, $l_{t,i}$. $l_{t,i}$ is one if the object is a risky object (i.e., in the positive class) and zero (i.e., in the negative class) otherwise. In real-world driving scenarios, positive and negative classes are imbalanced. Therefore, a weighted cross entropy loss is calculated for each video in the training dataset:
\begin{equation}
\mathcal{L} = -\sum_{t=1}^T\sum_{i=1}^M \left[w_p l_{t,i}\log(s_{t,i}) + w_n (1-l_{t,i}) \log(1-s_{t,i})\right],
\label{eq:loss}
\end{equation}
where $w_p$ and $w_n$ are the weights for the positive and negative classes, respectively. Losses of all the training videos are summed up to get the total loss for optimizing the learnable parameters.

%The goal of the training process is to fit all the parameters of the proposed model by backpropagating a differentiable loss. Given a video dataset which contains risky and non-risky objects. Each video of the dataset contains $T$ frames in total. At each time frame $t$, ground truth for $i$th object is denoted as  $l_{t,i}$ and predicted risk score is $s_{t,i}$. Since, in natural driving scenarios positive and negative classes are imbalanced, a weighted cross entropy loss is used. The weighted cross entropy loss for the $v$-th video is computed as the following equation: 

%Where, $w_p$ and $w_n$ are the class weights for positive and negative classes respectively. Losses from all the videos are summed up to get the total loss.  

\section{Dataset Development}
\label{sec:Dataset Development}
This study introduces a new dataset named Risky Object Localization (ROL) dataset and have made it available to the public \cite{rol_dataset_2022}. ROL has 1,000 video clips that each has 100 frames (i.e., 5-seconds long at the frequency 20 fps) and contains one or multiple risky traffic agents involved in an accident. The dataset provides the object-level, accident-level, and scene-level annotations for each video clip. The object-level annotation includes object classes (positive vs. negative, and traffic agent type), spatial annotation (i.e., where the traffic agents are present in frames), and temporal annotation of risky objects (when a risky object appears). The accident-level annotation includes accident type (ego involved vs non-involved), manner of collision, and temporal annotation (when it happens). The scene-level annotation includes road system, related to roadway, intersection related, weather, and lighting condition. 

TABLE \ref{tab:dataset_comparison} compares ROL with other traffic accident datasets. DAD, A3D and CCD datasets do not contain object-level and scene level annotations. Besides, these three datasets do not contain the manner of collision in their accident level annotation. DoTA dataset contains annotations at all levels. Yet, it does not provide the important accident time annotation. Besides, DoTA's annotation at the object level are different than the that of the proposed ROL dataset, as described in Section  \ref{sub:annotation}.

%This study introduced a publicly available dataset called Risky Object Localization (ROL) dataset. ROL has 1,000 video clips containing traffic accidents involving single and multiple risky traffic agents. 5 human annotators ensure the best quality of ROL dataset. Table -\ref{tab:dataset_comparison} compares ROL with other traffic accident datasets.

\begin{table}[t]
\renewcommand{\arraystretch}{1.3}
% \small\sf\centering
    \caption{Comparison between ROL dataset with other datasets. Information of other datasets are obtained from their released sources.}
    \label{tab:dataset_comparison}
    % \centering
    \resizebox{\linewidth}{!}{
    \begin{tabular}{l|r|r|c|c|c|c}
    \hline
    Dataset & \#videos & \# frames & fps & Accident & Object & Scene \\
    \hline
    DAD \cite{chan2016anticipating} & 620 & 62,000 & 20 & \checkmark & &  \\
    A3D \cite{yao2019egocentric} & 1,500 & 128,175 & 10 & \checkmark & &  \\
    CCD \cite{bao2020uncertainty} & 1,500 & 75,000 & 10 & \checkmark & &  \\
    DoTA \cite{yao2022dota} & 4,677 & 731,932 & 10 & \checkmark$^\ast$ &\checkmark$^{\star}$ & \checkmark  \\
    ROL (Ours) & 1,000 & 100,000 & 20 & \checkmark&\checkmark & \checkmark   \\
        \hline
%     \multicolumn{7}{l}{AA: accident annotation; OA: object annotation;}\\
%   \multicolumn{7}{l}{SA : scene annotation }\\
    \multicolumn{7}{l}{$\ast$ accident starting time is not available}\\
    \multicolumn{7}{l}{$\star$ different than the annotation method of ROL}\\
    \end{tabular}
    }
\end{table}

\subsection{Data Acquisition}

To develop the ROL dataset, videos are collected from the crowd sourced online platform YouTube using query terms like ``traffic accident" and ``road crash". A long video directly collected from the online platform usually contains many accidents. Histograms of image pixels in consecutive frames are similar for the same accident video segment but not for different video segments. Based on this fact, long videos are segregated into multiple video clips according to the similarity measurement of frame histograms and using 0.75 as the threshold. The resulting video clips may contain unnecessary video portions where the video starts too early before an accident, which have limited relevance to risky objects or the accident. Similarly, video clips have an unnecessarily post-accident portion. Therefore, the collected video clips are trimmed down to 5 seconds each with a frame rate of 20 frames per second (fps), resulting in 100 frames per video. Resolution of video frames are 1,080 $\times$ 720. The collected 1000 videos are then randomly divided into 800 training videos and 200 testing videos.

%To develop ROL, video clips are collected from crowd sourced online platform YouTube using query terms like "traffic accident" and "road crash". Diverse environmental attributes (eg. sunny, rainy, cloudy, and snow) and different lighting conditions (eg. day and night) are selected. This study avoided traffic accident videos where no risky object appears in the scene during the time of an accident. Therefore, traffic accidents such as loss of control of the ego vehicle where accidents happen without the involvement of any other object is not considered in this study.

%Directly collected long videos from online sources contain many accident video segments. Histograms of image pixels of consecutive frames of each video segment are similar but different for different video segments. Based on this fact, long videos are segregated into multiple video clips using histogram of image pixels with a threshold value of 0.75. The collected video clips may contain unnecessary video portions where the video starts too early before an accident, which does not have any relevance to the risky objects or the accident. Similarly, the end time may contain unnecessarily long portions of videos after the accident. Therefore, the collected video clips are trimmed down to 5 seconds each with a frame rate at 20 frames per second (fps), resulting in 100 frames per video. Resolution of the video frames are 1080 $\times$ 720. Collected 1000 videos are then randomly divided into 800 training videos and 200 testing videos. .

\subsection{Data Annotation}
\begin{figure}[htbp]
\centering
\includegraphics[width=3.1in]{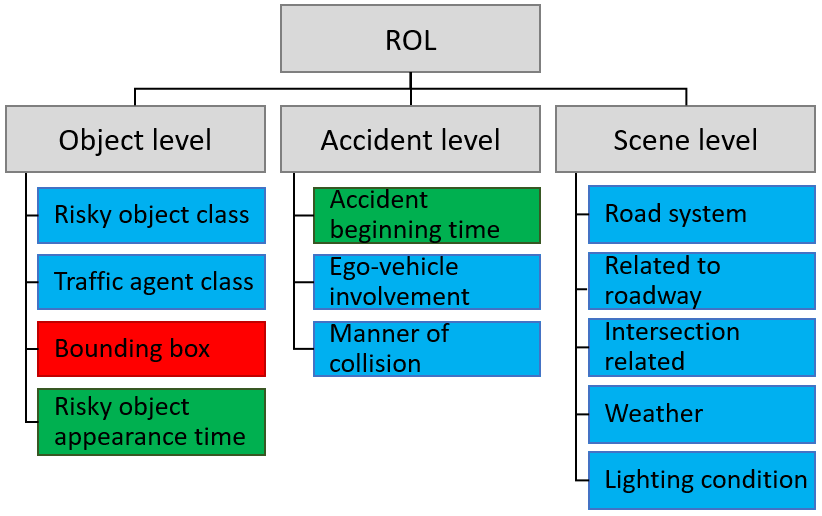}
\caption{Hierarchy of different annotation in ROL dataset. Blue, green, and red color indicate categorical, temporal, and spatial annotations, respectively.}
\label{fig_hierarchy}
\end{figure}

Fig-\ref{fig_hierarchy} shows the hierarchy of different annotations in the ROL dataset. This section further summarizes the temporal, spatial and categorical annotations of the dataset.
\label{sub:annotation}

%%%%%%%%%%%%%%%%%%%%%%%%%%%%%%%%%%%%%%%%%%%%%%%%%%%%%%%%%%%%%%%%%%%%%%%%%%%%%%%%%
%Figure: Sample images from ROL dataset
%%%%%%%%%%%%%%%%%%%%%%%%%%%%%%%%%%%%%%%%%%%%%%%%%%%%%%%%%%%%%%%%%%%%%%%%%%%%%%%%%
%%%%%%%%%%%%%%%%%%%%%%%%%%%%%%%%%%%%%%%%%%%%%%%%%%%%%%%%%%%%%%%%%%%%%%%%%%%%%%%%%

\subsubsection{Temporal Annotations}

Traffic agents in a video that will be involved in the accident in future frames are considered as risky objects. Both risky objects and the accident have temporal annotations. Since early detection of accident-involved objects is essential, this study annotates the time when a risky object appears in the driving scene for the first time, termed risky object appearance time. Unlike \cite{yao2022dota} that annotates the anomaly beginning time as the moment when annotators think an accident is inevitable, the approach to annotating risky object appearance time in this paper does not need annotators' subjective judgment. The accident beginning time is when a vehicle collides with another static or dynamic object. Accident beginning time is random in ROL dataset and the annotation is provided to set the anchor for calculating the earliness of detecting a risky object.

%The objects which will be involved in an accident in the future frames are considered as risky objects. Since early detection of accident is essential, this study annotated the time when a risky object appears in a scene for the first time, which is termed as risky object appearance time (RAT). These annotations do not need human judgment to decide when an object becomes risky. This study also annotated the accident beginning time (ABT) which is labeled at the time when a car crash happens. That is, when a vehicle collides with another static or dynamic object, it is considered as ABT. ABT is randomly placed in every video clip. The annotation of this study is different from   

\subsubsection{Spatial Annotation}

\begin{figure*}[htbp]
\centering
\includegraphics[width=7in]{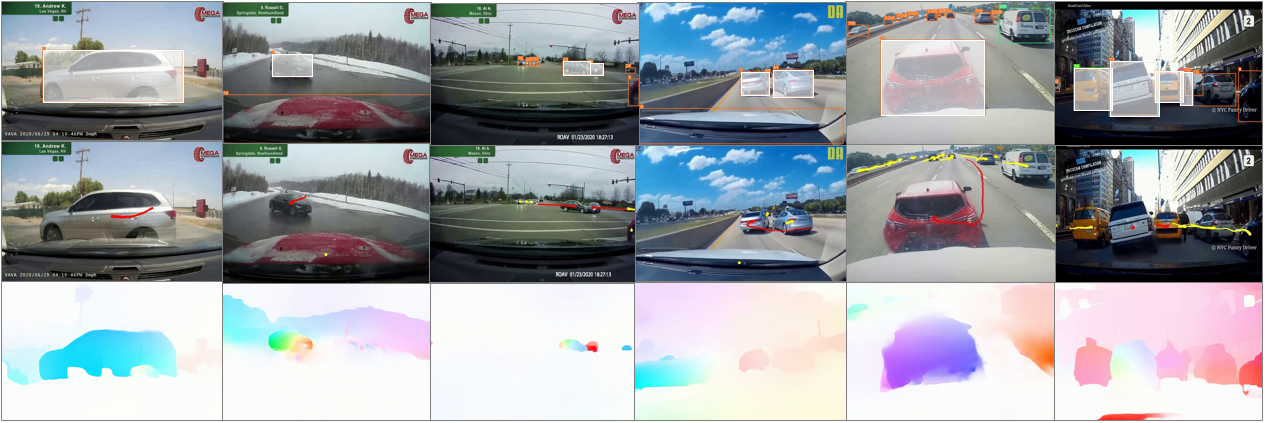}
\caption{Sample images from ROL Dataset}
\label{fig_sample}
\end{figure*}

ROL provides bounding boxes as the object-level spatial annotation in a semi-automated approach that keeps human annotators in the loop. The object detector YOLOv5 \cite{ge2021yolox} detects traffic agents and renders their bounding boxes. The multiobject tracker DeepSort \cite{wojke2017simple} further associates the same object over successive frames as a tracklet and provides a unique tracking ID. By watching videos overlaid with tracking IDs, human annotators identify the tracking IDs of risky objects. Solely relying on the object detector and the object tracker to provide the object-level spatial annotation could miss a few agents, thus lowering the quality of the dataset. In resolving this problem, the paper develops a click-to-select object tracker based on DaSiamRPN \cite{zhu2018distractor} as an additional intervention. The tool requires only a few mouse clicks by the human annotators to select missing or mis-tracked objects. Although it requires a little additional effort from annotators, the human intervention ensures the best quality of the dataset. Fig. \ref{fig_sample} illustrates six sample frames in ROL. The first row shows the spatial annotation of risky objects, in the form of white-shaded bounding boxes. The second row further visualizes object trajectories, with red curves for risky objects and yellow ones for non-risky objects. The third row are optical flow images of the sample frames.

%%%%%%%%%%%%%%%%%%%%%%%%%%%%%%%%%%%%%%%%%%%%%%%%%%%%%%%%%%%%%%%%%%%%%%%%%%%%%%%%%
%Figure: Sample images from ROL dataset
%%%%%%%%%%%%%%%%%%%%%%%%%%%%%%%%%%%%%%%%%%%%%%%%%%%%%%%%%%%%%%%%%%%%%%%%%%%%%%%%%

\subsubsection{Categorical Annotations}

Other annotations that ROL provides are about categorical variables, summarized in TABLE \ref{tab:category}. Traffic agent class is annotated automatically by YOLOv5, and others are annotated by human annotators. Annotators also rechecked the automatically annotated traffic agents and re-annotate miss-annotated positive traffic agents. Classes of accident-related and scene-related categorical variables generally follow the definition of FIRST \cite{collision_2022}.

%In addition to the spatial-temporal risky object annotation, ROL contains multiple important categorical labels in each video. For example, the dataset is split into ego-involved and non-ego involved videos. The dataset is also categorized in terms of accidents related to roadway types. The classes for this category are roadway, shoulder \& roadside, intersection, median, and others. These classes are defined based on the location where the accidents took place. ROL also contains the label for the manner of collisions as per the definition of \cite{collision_2022}, which are 1) angle collision: vehicles collide with each other in an angle, 2) sideswipe: two vehicles swipe each other from the side, 3) rear-end: one vehicle’s front crashes to the rear part of the other vehicle, 4) head-on: two vehicles crash in the front, and 5) others: vehicle crashes with a person, bicycle, and motorcycle, out-of-control crashes, running off-the-road, etc. Moreover, ROL contains labels for road type, weather, and lighting conditions. The labels with their statistics are summarized in table-\ref{tab:category}

\begin{table}[htbp]
\renewcommand{\arraystretch}{1.5}
    \caption{Summary of categorical variables and distributions}
    \label{tab:category}
    \centering
    \resizebox{\linewidth}{!}{
    \begin{tabular}{>{\arraybackslash}p{2.1cm}|>{\arraybackslash}p{4.9cm}}
        \hline
         Variable & Categories and distribution  \\
         \hline
         Traffic agent class & i) person (3.0\%), ii) car (84.1\%), iii) truck (11.2\%), iv) bus (0.9\%), v) motorcycle (0.5\%), vi) bicycle (0.2\%)  \\
         Risky object class & i) positive (21.3\%), ii) negative (78.7\%)\\
         Ego vehicle involvement  & i) ego involved (45.4\%), ii) ego non-involved (54.6\%)\\
         Road system & i) local (32.2\%), ii) arterial (36.8\%), iii) interstate (30.9\%) \\
         Related to roadway & i) roadway (52.5\%), ii) shoulder \& roadside (12.3\%), iii) median (2.4\%), iv) intersection related (29.2\%), v) others (3.6\%) \\
         Intersection related & i) no intersection (70.8\%) ii) 3 way (5.3\%), iii) 4 way (20.1\%), iv) others (3.8\%) \\
         Weather & i) clear (76.6\%), ii) cloudy (12.9\%), iii) rain (8.5\%), iv) snow (3.9\%)\\
         Lighting condition & i) day (81.3\%), ii) night (18.7\%) \\
         Manner of collision & i) angle (36.4\%), ii) sideswipe (11.6\%), iii) rear-end (19.2\%), iv) head-on (6.0\%), v) others (26.8\%) \\
        %  Manner of collision & 1) front-to-rear (19.2\%), 2) front-to-front (6.0\%), 3) angle (36.4\%), 4) sideswipe same direction (11.6\%) 5) others (26.8\%) \\
         \hline
    \end{tabular}
    }
\end{table}

\section{Results and Discussion}
\label{sec:results and Discussion}

Experiments are conducted to verify the effectiveness of the proposed AM-Net and the newly developed ROL dataset. Implementation details, evaluation metrics, and the results are discussed below.

\subsection{Implementation Details}
The proposed AM-Net is built using PyTorch \cite{paszke2019pytorch}. Model training and testing are performed using an Nvidia Tesla V100 GPU with 32GB of memory. All the input frames are resized to 224$\times$224 before feeding to the ResNet50 \cite{he2016deep} feature extractor. Feature vectors $\pmb{g}_{t,i}$ and $\pmb{o}_{t,i}$ are obtained by applying an average pooling operation to the output of the ResNet50 feature extractor, and their dimension ($D$) is 2,048. A fully-connected layer further reduces the dimension to 256 ($d$). The dimension of flow hidden representations ($N$) is 256 and that for bounding box hidden representations ($n$) is 32. A learning rate of 0.001 is used to train the AM-Net on the newly developed ROL dataset, and ReduceLROnPlateau is used as the learning rate scheduler. Adam optimizer is used to optimize the network for 30 epochs and the best model is selected. The positive to negative class ratio in ROL is $0.27: 1$. Therefore, the class weights for the negative class ($w_n$) and positive class ($w_p$) are selected as 0.27 and 1, respectively.

\subsection{Evaluation Metrics}

The evaluation of model performance focuses on two aspects: the correctness in localizing risky objects in videos and the earliness of risky object detection. To evaluate the correctness, this study uses Area under the Receiver Operating Characteristic Curve (AUC). AUC can measure the ability of AM-Net to differentiate risky and non-risky traffic agents.

To measure the earliness of the prediction, mean Time-to-Accident (mTTA) is used. Time-to-Accident (TTA) is defined as the first time when a riskiness score $s_{t,i}$ goes across a threshold value $\bar{s}$. That is,
\begin{equation}
    \text{TTA}=\max\{\tau-t|s_{t,i}\ge \bar{s}, 0\le t\le \tau, \forall\, i\},
\end{equation}
where $\tau$ is the accident beginning time. TTA is dependent of the selection of a threshold value $\bar{s}$. mTTA, the average of TTA values at different threshold values, is calculated as an earliness metrics independent of the threshold value selection.

\subsection{Evaluation of the Proposed Model Architecture}

An ablation study is conducted to evaluate the contribution of different components of AM-Net, which compares the correctness and earliness metrics of AM-Net and nine variants of it. Those ten models are trained and tested on ROL, and results are summarized in TABLE \ref{tab:ablation}.

%In order to verify the effectiveness of the proposed framework and the developed ROL dataset, multiple experiments were conducted on ROL dataset using different variants of the proposed method. These variants illustrates the contribution of different components of the design choice.Table \ref{tab:ablation} summarizes the correctness and earliness attained with the different variants of the proposed method. 

\begin{table}[]
\renewcommand{\arraystretch}{1.3}
\centering
\caption{Ablation study on the ROL dataset.}
    \label{tab:ablation}
\resizebox{\linewidth}{!}{
\begin{tabular}{l|cc|c|cc|c|cc}

\hline
\multirow{2}{*}{Model} & \multicolumn{2}{c|}{RGB}            & \multicolumn{1}{c|}{\multirow{2}{*}{Bbox}} & \multicolumn{2}{c|}{Flow}  & \multirow{2}{*}{Att.} & \multicolumn{1}{c}{AUC}                       & mTTA                    \\ 
\cline{2-3} \cline{5-6}
                    & O & \multicolumn{1}{c|}{F} & \multicolumn{1}{c|}{}                      & O & \multicolumn{1}{c|}{F} & &\multicolumn{1}{c}{(\%)} &  \multicolumn{1}{c}{(s)} \\
                    \hline
1&\checkmark&\checkmark&&&&&80.13&2.30\\
2&&&\checkmark&&&&82.01&2.24\\
3&&&&\checkmark&\checkmark&&81.49&2.09\\
4&\checkmark&\checkmark&\checkmark&&&&83.46&2.16\\
5&\checkmark&\checkmark&&\checkmark&\checkmark&&82.43&2.05\\
6&\checkmark&\checkmark&\checkmark&\checkmark&\checkmark&&85.11&2.08\\
7&\checkmark&&\checkmark&\checkmark&\checkmark&&85.20&2.01\\
8&&&\checkmark&\checkmark&\checkmark&&85.47&2.05\\
9&&&\checkmark&\checkmark&&&84.11&1.87\\
10&&&\checkmark&\checkmark&\checkmark&\checkmark&85.73&2.34\\
\hline
\multicolumn{9}{l}{O: object level feature; F: frame level feature; Att: Attention;} \\
\multicolumn{9}{l}{Bbox: Bounding box feature}\\
\end{tabular}
}
\end{table}

% \begin{table}[]
% \renewcommand{\arraystretch}{1.3}
% \centering
% \caption{Ablation study on the ROL dataset.}
%     \label{tab:ablation}
% \resizebox{\linewidth}{!}{
% \begin{tabular}{l|cc|c|cc|c|ccc}

% \hline
% \multirow{2}{*}{Model} & \multicolumn{2}{c|}{RGB}            & \multicolumn{1}{c|}{\multirow{2}{*}{Bbox}} & \multicolumn{2}{c|}{Flow}  & \multirow{2}{*}{Att.} & \multicolumn{1}{c}{AUC}  & AP                       & mTTA                    \\ 
% \cline{2-3} \cline{5-6}
%                     & O & \multicolumn{1}{c|}{F} & \multicolumn{1}{c|}{}                      & O & \multicolumn{1}{c|}{F} & &\multicolumn{1}{c}{(\%)} & \multicolumn{1}{c}{(\%)} & \multicolumn{1}{c}{(s)} \\
%                     \hline
% 1&\checkmark&\checkmark&&&&&80.13&47.84&2.30\\
% 2&&&\checkmark&&&&82.01&49.91&2.24\\
% 3&&&&\checkmark&\checkmark&&81.49&54.77&2.09\\
% 4&\checkmark&\checkmark&\checkmark&&&&83.46&51.86&2.16\\
% 5&\checkmark&\checkmark&&\checkmark&\checkmark&&82.43&54.30&2.05\\
% 6&\checkmark&\checkmark&\checkmark&\checkmark&\checkmark&&85.11&57.69&2.08\\
% 7&\checkmark&&\checkmark&\checkmark&\checkmark&&85.20&59.24&2.01\\
% 8&&&\checkmark&\checkmark&\checkmark&&85.47&60.18&2.05\\
% 9&&&\checkmark&\checkmark&&&84.11&57.55&1.87\\
% 10&&&\checkmark&\checkmark&\checkmark&\checkmark&85.73&61.76&2.34\\
% \hline
% \multicolumn{10}{l}{O: object level feature; F: frame level feature; Att: Attention;} \\
% \multicolumn{10}{l}{Bbox: Bounding box feature}\\
% \end{tabular}
% }
% \end{table}

{
Model \#1 from the table solely uses the object-level and frame-level appearance features that can be directly extracted from the RGB images as the input for risky object localization. This model achieves 80.13\% AUC with 2.30 second mTTA. This performance sets a benchmark for evaluating the proposed AM-Net (i.e., model \#10) and its variants (models \#2$\sim$9). Model \#2 uses the bounding box feature as the single input, which increases AUC to 82.13\%  but reduces mTTA slightly for 0.06 second. This suggests that the bounding box feature is better than appearance features to improve the correctness of detecting risky objects in traffic videos. Changes in bounding box location and scales are more discriminative than appearance features for risky object detection. Similarly, Model \#3 uses flow features as the input. The comparison between models \#1 and \#3 shows flow features in standalone can improve AUC for 1.36\%, but it reduces mTTA for 0.21 second, in detecting risky objects. In short, the bounding box feature and flow features can respectively improve the correctness, but lower the earliness, in predicting risky objects.
}

{
The fusion of bounding box feature with appearance features (model \#4) achieves a higher AUC than using any individual feature type (models \#1 and \#2), but not longer mTTA. The similar conclusion can be made by comparing the fusion of flow features with appearance features (model \#5) to the use of individual feature type (models \#1 and \#2). Fusing those three types of features together (model \#6) effectively brings AUC to 85.11\%. Models \#4$\sim$\#6 confirms the effectiveness of feature fusion.  
}

{Model \#7 removes the frame-level appearance feature input from model \#6, which increases the AUC slightly, for 0.09\%, but sacrifices the mTTA for 0.07 second. Model\#8 removes both the frame-level and object-level appearance features from model \#6, resulting an additional increase in AUC. Model \#9 removes the frame-level flow feature of model \#8, resulting in lower AUC and shorter mTTA. The frame-level flow feature captures the motion of the traffic scene due to the ego-vehicle movement and the motion of other objects in the same frame, which are important clue for identifying risky objects in the frame. The comparison of those four models (\#6$\sim$\#9) suggests the fusion of bounding box feature and flow features (both the object level and the frame level) is the most suitable design of feature fusion. Appearance features, while are effective in other tasks, may lower the ability to differentiate objects by their riskiness classes. 
}

{Finally, model \#10 (i.e., the proposed AM-Net) adds the proposed attention module to model \#8, which achieves the best performance, 85.73\% AUC and 2.34 second mTTA. The attention module not only effectively addresses the limitation of bounding box feature and flow features in detecting risky objects earlier but further increases the correctness.
}

\subsection{Comparison to the State-of-the-Art Models}

{
This study compares the AM-Net (both without and with the attention module) trained on ROL (i.e., models \#8 and \#10 in TABLE \ref{tab:ablation}) to existing models \cite{hasan2016learning, chong2017abnormal, liu2018future, yao2022dota} on the testing dataset of DoTA \cite{yao2022dota} from the perspective of video anomaly detection. That is, they are compared on the metric of frame-level AUC. The earliness of detection is skipped in this comparison because DoTA testing dataset does not have the annotation of accident beginning time. TABLE \ref{tab:sota} summarizes results of the comparative study. The frame-level AUC values of models \cite{hasan2016learning, chong2017abnormal, liu2018future, yao2022dota} are provided by \cite{yao2022dota}. This paper computes the frame-level AUC for the proposed AM-Net. To compute the frame-level AUC with the output of AM-Net, the highest riskiness score of any detected objects in a frame, $\max_{i}\{s_{t,i}\}$, is considered as the riskiness score for the frame. A frame containing a risky object is a positive sample. The Receiver Operating
Characteristic (ROC) curve is attained by calculating the true positive rate and false positive rate at various detecting threshold values. Then, the frame-level AUC is calculated accordingly. 
}

\begin{table}[htbp]
\renewcommand{\arraystretch}{1.3}
    \caption{Comparison of the proposed model with existing methods on DoTA \cite{yao2022dota} testing dataset for video anomaly detection}
    \label{tab:sota}
    \centering
    \resizebox{\linewidth}{!}{%
    \begin{tabular}{llr}
        \hline
         Method & Features & Frame-level AUC(\%)  \\
         \hline
         ConvAE \cite{hasan2016learning} & Flow (F) & 66.3  \\
         ConvLSTMAE \cite{chong2017abnormal} & Flow (F) & 62.5 \\
         FFP \cite{liu2018future} & RGB(F) & 67.5\\
         FOL \cite{yao2022dota} & RGB(F) + Bbox + Flow(O) + Ego & 73.0\\
         \hline
         %\textbf{AMS-Net (Ours)} & RGB (O+F) + Bbox + Flow (O+F) & 75.7 \\
         %& RGB (O) + Bbox + Flow (O+F) & 78.8 \\
         \textbf{AM-Net (Ours)} &Bbox + Flow (O+F) & \textbf{79.3} \\
         &Bbox + Flow (O+F) + Attn. & 75.7\\ 
        %  &Bbox + Flow (O+F) + Attention & 75.7 \\
         \hline
    \end{tabular}
    }
\end{table}

{
TABLE \ref{tab:sota} shows that ConvAE  and ConvLSTMAE with the frame-level flow feature as the only input achieve 66.3\% and 62.5\% AUC, respectively. FFP, which uses the frame-level appearance feature, achieves 67.5\% AUC. FOL effectively boosts up the performance to 73.0\% by fusing the bounding box, ego motion, and object-level optical flow feature with the frame-level appearance feature. The AM-Net without the attention module achieved 79.3\%, about 6.3\% higher than FOL attains. The improvement of AM-Net above FOL can be associated with the choice of features and the method that the video anomaly detection is based upon. The comparison also signifies that the bounding box and flow features are the most important features to capture the spatio-temporal pattern of risky objects. When testing the AM-Net with the attention module on the DoTA testing dataset, a 75.71\% frame-level AUC is achieved. Adding the attention module to AM-Net does not attain the best performance in this test because AM-Net is trained on ROL that has 20 fps but tested on DoTA that has only 10 fps. The temporal dynamics of the spatial relationship between different objects are hampered when a large portion of a video sequence is skipped. Therefore, when the frame frequency of testing videos is low, the spatial attention learnt from a training dataset in higher frame frequency has a negative effect. It should be noted that the performance of AM-Net can be increased further if it was both trained and tested on the DoTA dataset. 
}

\subsection{Performance by Manner of Collision}

{The spatio-temporal pattern of risky objects varies with the manner of collision. To reveal how well the proposed AM-Net is performing in localizing risky objects involved in different manners of collision, this study calculates the AUC and mTTA by the manner of collision. The testing dataset of ROL contains 34.0\% angle, 18.5\% sideswipe, 10.0\% rear-end, 5.5\% head-on, and 32.0\% other types of collision videos.
}

%Risky objects from different manner of collisions have different spatio-temporal patterns. It is important to understand how well the network performs in different situations. This study evaluates the performance of the proposed network regarding the manner of collision. To evaluate this, the ROL test dataset is decomposed based on the manner of collision. The test dataset contains 34\% angle, 18.5\% sideswipe, 10\% rear-end, 5.5\% head-on, and 32\% others type of collision videos. AUC, AP and mTTA are computed on different types of collisions to get the results for individual manner of collision. Table \ref{tab:manner} shows the performance with respect to the manner of collision. 

\begin{table}[htbp]
% \small\sf\centering
\centering
\renewcommand{\arraystretch}{1.3}
\caption{Performance by manner of collision}
    \label{tab:manner}

\begin{tabular}{l|cc}
\hline
Manner of  collision & AUC (\%)       & mTTA (s) \\ \hline
Angle                & 84.71  &  2.31 \\
Sideswipe            & 88.95  & 2.13 \\
Rear-end        & 90.00  & 1.97 \\
Head-on       & 76.32 & 2.14 \\
Others               & 82.03  & 2.70 \\ \hline
\end{tabular}
\end{table}

%%%%%%%%%%%%%%%%%%%%%%%%%%%%%%%%%%%%%%%%%%%%%%%%%%%%%%%%%%%%%%%%%%%%%%%%%%%%%%%%%
\begin{figure*}[!bt]
\centering
\includegraphics[width=6in]{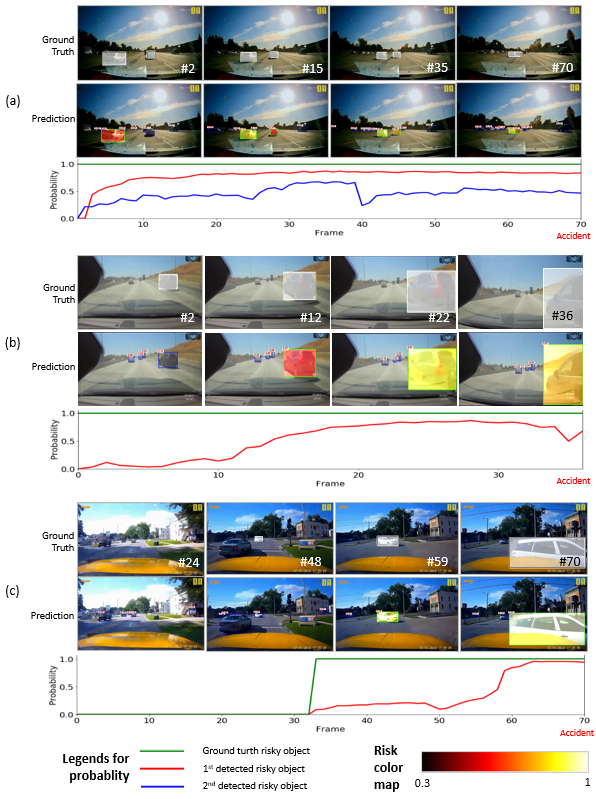}
\caption{Examples of risky object localization.%: a) examples with two risky objects, b) example where a single risky object involved with the ego vehicle, and c) example of ego vehicle involved accident where risky object appears after 32 frame. Ground truth risky object is shaded with white bounding boxes, the predicted risky objects are shaded with color as per the color scheme shown in risk color map based on the intensity of probability. The green line shows the ground truth's probability. Red and blue curvature show the predicted probability along the timeline. The end point of the timeline is the accident beginning time.
}
\label{fig_qual}
\end{figure*}
%%%%%%%%%%%%%%%%%%%%%%%%%%%%%%%%%%%%%%%%%%%%%%%%%%%%%%%%%%%%%%%%%%%%%%%%%%%%%%%%%

% \begin{table}[htbp]
% % \small\sf\centering
% \centering
% \renewcommand{\arraystretch}{1.3}
% \caption{Performance by manner of collision}
%     \label{tab:manner}

% \begin{tabular}{l|ccc}
% \hline
% Manner of  collision & AUC (\%)   & AP (\%)    & mTTA (s) \\ \hline
% Angle                & 84.71 & 61.34 &  2.31 \\
% Sideswipe            & 88.95 & 69.98 & 2.13 \\
% Rear-end        & 90.00 & 71.19 & 1.97 \\
% Head-on       & 76.32 & 46.67 & 2.14 \\
% Others               & 82.03 & 56.49 & 2.70 \\ \hline
% \end{tabular}
% \end{table}

% \begin{table}[htbp]
% \small\sf\centering
% \renewcommand{\arraystretch}{1.3}
% \caption{Performance with respect to manner of collision}
%     \label{tab:manner}

% \begin{tabular}{l|ccc}
% \hline
% Manner of  collision & AUC (\%)   & AP (\%)    & mTTA (s) \\ \hline
% Angle                & 87.33 & 65.27 & 1.95 \\
% Sideswipe            & 87.54 & 66.12 & 1.96 \\
% Rear-end        & 88.34 & 62.66 & 1.82 \\
% Head-on       & 78.57 & 40.51 & 2.00 \\
% Others               & 82.02 & 53.40 & 2.69 \\ \hline
% \end{tabular}
% \end{table}

The highest AUC that AM-Net achieves is 90.00\%, in predicting risky objects in the rear-end collision. Risky objects in this manner of collision are in front and seem less challenging to localize. While localizing risky objects in the sideswipe collision is more difficult than in the rear-end collision, AM-Net still attains 88.95\% AUC. The AUC pertaining to the angle collision is 84.71\%, and the large variability of videos with this manner of collision is one possible reason that correctly localizing risky objects becomes more difficult. AM-Net only achieves 76.32\% AUC for the head-on collision. Data for head-on collision is very rare in natural driving videos. Hence, the small portion of training videos in this collision type is the major reason for the relatively weaker performance. On testing videos of other manners of collision, AM-Net achieves 82.03\% AUC, mainly due to the large variability of the spatio-temporal pattern of risky-objects.

If other manners of collisions are not counted, AM-Net is the fastest in localizing risky objects in the angle collision, using 2.31 second on average. AM-Net can predict risky objects in the sideswipe and head-on collision in about 2.13$\sim$2.14 second, but slightly less than 2 second in the rear-end collision. More or less, the relatively short mTTA in localizing risky objects in the rear-end collision pertains to the relatively close distance of those objects to the ego-vehicle in this manner of collision.

The experimentation with respect to different manners of collision confirms that AM-Net can achieve promising performance in localizing risky objects, particularly in angle, sideswipe, and rear-end collision types. As per NHTSA report \cite{collision_2022}, 66\% of reported collisions with body injury or property damage pertain to these three manners of collision. Correctly localizing risky objects ahead of time would allow for taking preventive actions to avoid the accident.

\subsection{Qualitative Evaluation}
% Figure \ref{fig_qual} illustrates three examples of risky object localization results of the AM-Net with samples from ROL Dataset. In each example, ground truth risky objects in the sample frames of the video are shown on the top row with shaded white bounding boxes, middle row shows the predicted risk score for all the detected objects. All the detected objects are localized using bounding boxes. For the illustration purpose, a threshold value of 0.3 is set to define an object as a risky object. Risky objects are shaded with color. A color scheme of dark reddish black to yellow to white is used to illustrate the intensity of the risk score. A complete dark reddish black shaded bounding box indicates a risk score of 0.3 and a complete white shaded bounding box indicates a risk score of 1. The bottom row of each example shows the predicted riskiness score corresponding to the ground truth objects along the timeline with colored curve. The green curve corresponds to the ground truth objects. The ground truth objects risk score is 1. The red curve shows the predicted riskiness score for the first risky object and the blue curve shows for the 2nd risky object if there is any second risky object. The timeline is shown till the actual collision happens. That is the last point of the timeline along the horizontal axis is the accident beginning time.
%Figure: Qualitative example

Fig. \ref{fig_qual} illustrates three examples in the ROL dataset to demonstrate the risky object localization result of AM-Net. In each example, the true risky objects are highlighted by white-shaded bounding boxes. The riskiness score of true risky objects in a frame is one as long as it appears in that frame, which is indicated by the green curve in the bottom row. Risky objects localized by AM-Net are overlaid with colored bounding boxes. For the illustration purpose, a threshold value of 0.3 is set to determine if a predicted object is a risky object. Therefore, the color scheme for bounding boxes shown in Fig. \ref{fig_qual}, which is a reddish black color to a yellowish white color, is mapped to the range [0.3, 1].  The riskiness score of riskiness objects are presented as curves in colors other than green.

%
%Figure \ref{fig_qual} illustrates three examples of risky object localization results produced by AM-Net with samples from the ROL Dataset. In each example ground truth risky objects (shaded white) are shown in the top row, predicted risky objects (shaded with color) are in the middle row and the risk probability score of those detected risky objects are shown in the bottom row. For the illustration purpose, a threshold value of 0.3 is set to define a predicted object as a risky object. Therefore the color scheme shown in the Figure \ref{fig_qual}, which is a reddish black color to a yellowish white color is mapped to [0.3, 1].  

The video of example (a) contains two risky objects, which are two vehicles far away from the ego vehicle. They appear in the video from the beginning and, thus, the ground truth riskiness score (i.e., the green line) is one throughout the timeline of the video sequence. The video contains a risky situation where two vehicles are approaching each other from the side, resulting in a collision starting at frame \#70. The accident scene is complex because the two risky objects are very small due to the far distance from the ego-vehicle. AM-Net successfully assigns the highest riskiness scores to those two objects from a very early stage. The color of the first risky object's bounding box (in the middle row of the example) quickly turns from red to yellow, and the riskiness score (the red curve in the bottom row of the example) increases gradually. For the second risky object, its riskiness score (the blue curve) is a little lower than that of the first object. It is mainly because of the relatively lower  velocity and relatively smaller size of the second vehicle than the first one. After reaching the 0.3 threshold value, the network successfully maintains the riskiness score above the threshold value across the timeline.

%Example a) contains two risky objects where both the objects are far away from the camera. The risky objects appeared in the scene at the beginning of the video at \#0-th frame. Therefore the ground truth risk probability (i.e. green line) is 1 throughout the timeline. The scene contains a risky situation where two vehicles are approaching to each other from the side to have a collision at \# 70-th frame. The scene is very complex because both the objects are very small due to the far distance from the camera. The proposed network successfully assigned highest riskiness score to those two objects from the very early stage of the timeline. For risky object one, the color of the bounding box turned from red to yellow. That is the risk probability gradually increased, which is also reflected in the probability curve (i.e. red curve) in the bottom of the example. For the second risky object the riskiness score (blue curve) is little lower than the first object. It is mainly because of the low relative velocity of the second vehicle than the first vehicle and relatively smaller size than the first one. After reaching the threshold the network successfully maintains the riskiness score above the threshold value across the timeline.

The video of example (b) is an ego vehicle-involved accident. One vehicle (the risky object) cuts into the direct lane of the ego vehicle from the side, resulting in a collision at frame \#36. In this example, the risky object appears in the video from the beginning. AM-Net assigns the highest riskiness score to the correct vehicle in the traffic.

%Example b) contains a sample, where the ego vehicle is involved with an accident. In this example, a vehicle cuts into the direct way of the ego vehicle from the side to have a collision at frame \# 36. Here also the risky vehicle appeared in the scene at frame \# 0. The proposed method successfully assigned highest riskiness score to the correct vehicle. 

In the video of example (c), the risky object is not there until frame \#33 when the vehicle is coming to the front of the ego vehicle from the opposite direction, resulting in an angle collision at frame \#70. Although the risky object appears for just a short period, AM-Net does not predict any false positive detection. Instead, it successfully assigns the highest riskiness score to the correct vehicle.

%Example c) shows a sample, where the risky object did not appear in the first 32 frame. The risky vehicle appeared in the frame \# 33, where the vehicle was coming to the direct way of the ego vehicle from the opposite direction to have an angle collision with the ego vehicle at frame \# 70. Although, risky object did not appear for a long period, the proposed network did not predict any false positive detection and the proposed method successfully assigned the highest riskiness score to the correct vehicle. 
% \vfill

\section{Conclusion}
\label{sec:Conclusion}

This paper presented a novel deep learning framework named AM-Net for localizing risky traffic agents from dashcam videos. AM-Net extracts the spatial information (i.e., bounding boxes) and motion information (i.e., optical flows) of traffic agents in the driving scene and learns those agents' spatio-temporal features from such information in successive frames. The network effectively localizes risky agents in the driving videos by fusing the multistream features at the object and frame levels and learning to differentiate its attention to different agents. A thorough ablation study not only justifies the rationale for selecting bounding boxes and optical flows as the input but verifies the effectiveness of the proposed feature fusion and attention mechanism. Moreover, experimental evaluation confirms that AM-Net has exceeded the state-of-the-art. The paper also introduced ROL, a benchmark dataset containing object-level, accident-level, and scene-level annotations for the task of risky object localization in dashcam videos. Besides, the rich annotations of ROL capture multiple communities' interests and can fuel multidisciplinary research on intelligent systems for transportation safety enhancement.

%This paper presented a novel AM-Net for localizing risky traffic agents from dashcam videos. The proposed network effectively learns spatial-temporal relationships between consecutive video frames to localize the risky traffic agents. The network also learns to attend to the important traffic agents relevant to a future risk with the help of the proposed attention module. Besides, AM-Net uses feature fusion among different object-level and frame-level features that were extracted from the input RGB video stream to boost up the performance. According to this study, the bounding box and flow features are more effective at predicting risky objects than the raw RGB features. A thorough ablation study demonstrates the effectiveness of the proposed feature fusion and attention mechanism. Moreover, experimental evaluation confirms that the proposed method has exceeded state-of-the-art performance. This study also introduced ROL, a benchmark dataset containing object-level, accident-level, and scene-level annotations for the task of risky object localization in dashcam videos. Besides, rich annotations of ROL will accelerate the convergence of multiple research disciplines of traffic safety enhancement.

This research builds a foundation for exploring advanced opportunities for improving the performance of risky object localization, such as multimodal sensor fusion, multi-camera perception, and Vehicular Ad-hoc Networks (VANET). For example, camera-LiDAR fusion allows for extending this paper by building the model upon reliable 3D object detection and tracking. 3D visual perception based on a multi-camera system has become a critical task of autonomous driving. Localizing dangerous traffic agents 360$^\circ$ around the ego vehicle will vastly improve safety. Integrating the AM-Net into the Vehicular Ad hoc Network (VANET) environment allows for relaying the information of risky traffic agents to connected vehicles and infrastructure to minimize the impact of a potential accident on the traffic.

While AM-Net has achieved promising performance in localizing risky traffic agents and anticipating traffic accidents, its limitations are present and should be addressed in future research. Data scarcity limits the early localization of the risky objects from head-on collisions in this study. Incorporating the prior knowledge from the dataset into the model learning to distribute more attention to the front-facing vehicles approaching the ego vehicle may address this issue. Another limitation is that this study considers only six types of traffic agents. The model can further include other traffic agents, especially animals in specific areas and road debris.

\section*{Acknowledgment}
%Qin and Karim receive support from National Science Foundation (NSF) through the grant ECCS-\#2026357. Yin and Karim receive support from NSF through ECCS-\#2025929.
It will be added after the manuscript is accepted for publication.
\ifCLASSOPTIONcaptionsoff
  \newpage
\fi



\bibliographystyle{IEEEtran}
\bibliography{ref}

%\vfill


\end{document}